\documentclass[sigconf,screen]{acmart}
\AtBeginDocument{%
  \providecommand\BibTeX{{%
    \normalfont B\kern-0.5em{\scshape i\kern-0.25em b}\kern-0.8em\TeX}}}

\usepackage{graphics}
\usepackage{subfigure}
\copyrightyear{2022}
\acmYear{2022}
\setcopyright{acmlicensed}\acmConference[MM '22]{Proceedings of the 30th ACM International Conference on Multimedia}{October 10--14, 2022}{Lisboa, Portugal}
\acmBooktitle{Proceedings of the 30th ACM International Conference on Multimedia (MM '22), October 10--14, 2022, Lisboa, Portugal}
\acmPrice{15.00}
\acmDOI{10.1145/3503161.3547996}
\acmISBN{978-1-4503-9203-7/22/10}

\begin{document}

\title{Video-Guided Curriculum Learning for Spoken Video Grounding}

\author{Yan Xia}
\affiliation{%
  \institution{Zhejiang University}
  \city{Hangzhou}
  \country{China}}
\email{xiayan.zju@gmail.com}

\author{Zhou Zhao}
\authornote{Corresponding author.}
\affiliation{%
  \institution{Zhejiang University}
  \city{Hangzhou}
  \country{China}}
\email{zhaozhou@zju.edu.cn}

\author{Shangwei Ye}
\affiliation{%
  \institution{Zhejiang University}
  \city{Hangzhou}
  \country{China}}
\email{yeshangwei@zju.edu.cn}

\author{Yang Zhao}
\affiliation{%
  \institution{Zhejiang University}
  \city{Hangzhou}
  \country{China}}
\email{awalk@zju.edu.cn}

\author{Haoyuan Li}
\affiliation{%
  \institution{Zhejiang University}
  \city{Hangzhou}
  \country{China}}
\email{lihaoyuan@zju.edu.cn}

\author{Yi Ren}
\affiliation{%
  \institution{Zhejiang University}
  \city{Hangzhou}
  \country{China}}
\email{rayeren613@gmail.com}


\renewcommand{\shortauthors}{Yan Xia et al.}
\begin{abstract}
  In this paper, we introduce a new task, spoken video grounding (SVG), which aims to localize the desired video fragments from spoken language descriptions. Compared with using text, employing audio requires the model to directly exploit the useful phonemes and syllables related to the video from raw speech. Moreover, we randomly add environmental noises to this speech audio, further increasing the difficulty of this task and better simulating real applications. To rectify the discriminative phonemes and extract video-related information from noisy audio, we develop a novel video-guided curriculum learning (VGCL) during the audio pre-training process, which can make use of the vital visual perceptions to help understand the spoken language and suppress the external noise. Considering during inference the model can not obtain ground truth video segments, we design a curriculum strategy that gradually shifts the input video from the ground truth to the entire video content during pre-training. Finally, the model can learn how to extract critical visual information from the entire video clip to help understand the spoken language. In addition, we collect the first large-scale spoken video grounding dataset based on ActivityNet, which is named as ActivityNet Speech dataset. Extensive experiments demonstrate our proposed video-guided curriculum learning can facilitate the pre-training process to obtain a mutual audio encoder, significantly promoting the performance of spoken video grounding tasks. Moreover, we prove that in the case of noisy sound, our model outperforms the method that grounding video with ASR transcripts, further demonstrating the effectiveness of our curriculum strategy. The code is available at \url{https://github.com/marmot-xy/Spoken-Video-Grounding}. 
\end{abstract}

\ccsdesc[500]{Computing methodologies~Activity recognition and understanding}
\keywords{datasets, contrastive learning, curriculum learning, video grounding}


\maketitle

\section{Introduction}
\label{sec:intro}

With the explosive growth of videos on the Internet, systematically understanding video contents and quickly localizing the desired video fragments through a query has become increasingly important. Existing datasets \cite{DBLP:conf/iccv/GaoSYN17, DBLP:conf/cvpr/HeilbronEGN15} and approaches \cite{DBLP:conf/cvpr/MunCH20, DBLP:conf/aaai/HeZHLLW19, zhang2020span} use text descriptions to locate the corresponding segments in the video, which have achieved promising results. However, with the development of the Automatic Speech Recognition (ASR) and Text To Speech (TTS), speech is becoming an essential medium for Human-to-machine interaction. Thus in this paper, to investigate whether unsegmented spoken language can highlight corresponding segments in unconstrained videos, we propose a new task named spoken video grounding (SVG), which exploits raw speech audio as input to localize the starting and ending video frames. Furthermore, in real-world scenarios, when people want to use their speech voice to interact with their smart device, i.e., mobile phones, or service robots, it is hard to guarantee their voice input will not be disturbed by any external noise. These noises can come from various sources, such as the sound of cars on the road and noisy people in the hall. Therefore, to better simulate the real-world application, we also randomly mix environment noises with the speech in our dataset. We are dedicated to suppressing this environment noise and localizing the video from speech voice more robustly. 

Previous works have studied how to combine audio and vision together. Harwath et al. \cite{DBLP:conf/iclr/HarwathHG20} and \cite{DBLP:journals/ijcv/HarwathRSCTG20} utilize speech signals to highlight the relevant image regions. They think the process is similar to how babies recognize words and visual objects \cite{gomez2000infant}.
However, spoken language is more related to visual events rather than still images. Thus \cite{DBLP:journals/corr/abs-2006-09199} and \cite{DBLP:conf/cvpr/BoggustAJHTFGZ019} use instructional videos to learn the semantic connection between raw speech and visual entities without any ASR transcripts. However, solely using instructional videos is limited to video events and spoken words. Therefore, in this paper, we bring up a novel spoken language video grounding task based on our new proposed dataset named ActivityNet Speech. The ActivityNet Speech dataset is extended from ActivityNet, which is open-domain and contains various activity types. The new task demonstrates that text annotations are not necessary to understand visual environments, machine can obtain meaningful linguistic abstractions like phonemes and syllables directly from speech signals.

\begin{figure}[tb] 
 \center{\includegraphics[width=9cm]  {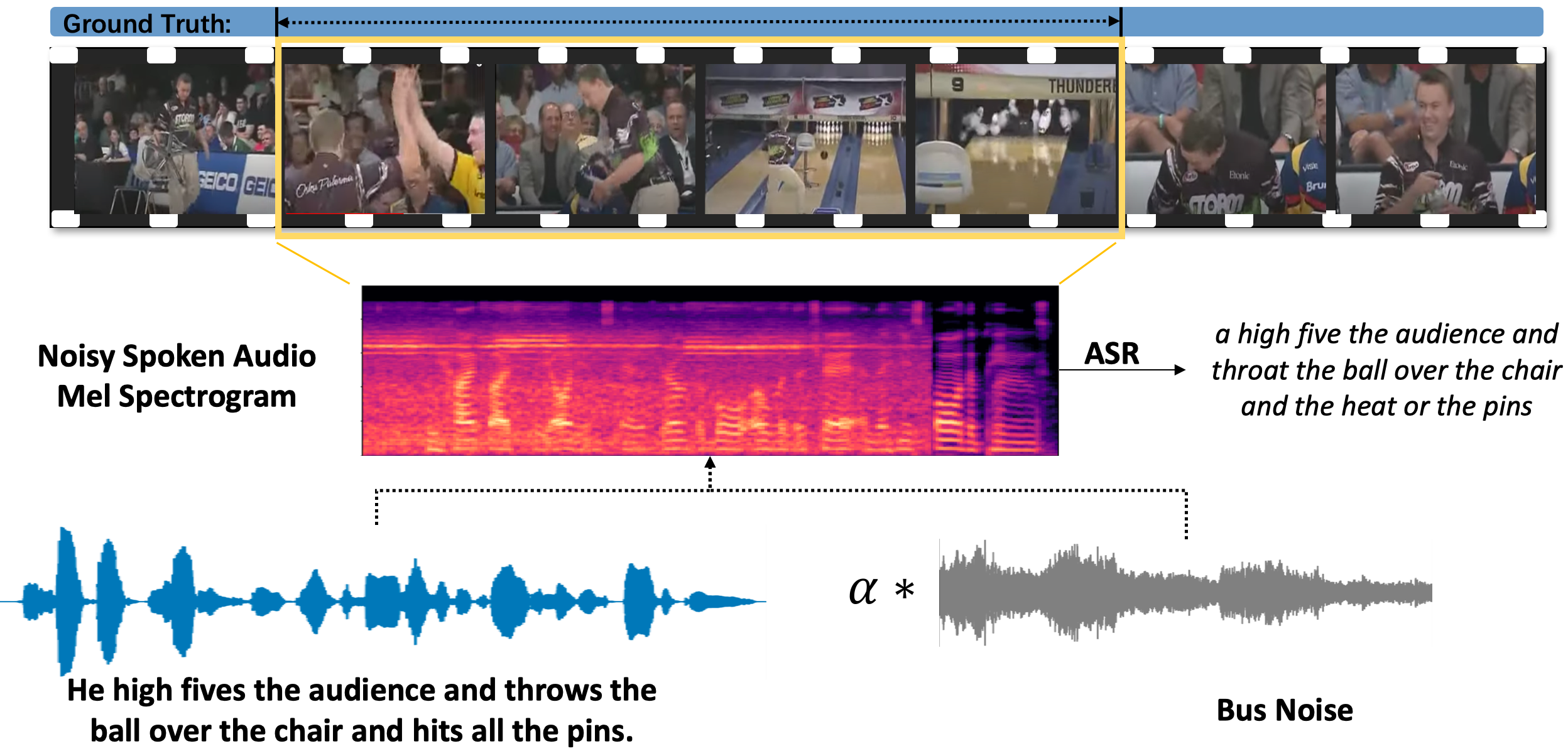}} 
 \caption{\label{fig1} The overall of our proposed task. We directly use the noisy spoken audio to localize the corresponding video segment. The ASR transcripts show that there are lots of missing or wrong words due to the noise, which will mislead the grounding model if directly using the transcript results.}
 \vspace{-1.0em}
\end{figure}
Our proposed task is closely related to text-to-video grounding. However, three limitations prevent it from being directly used in the spoken-to-video grounding task, specifically: 1) Utilizing ASR to recognize raw speech imposes several errors due to background sounds and accents. Further, 98\% of languages are not available for ASR transcripts \cite{DBLP:conf/interspeech/PrasadERM19}. 2) ASR converts continuous unsegmented speech signals into discrete text signals. Introducing natural noise makes the words recognized by ASR have similar pronunciations but different semantics, i.e. 'adapt' and 'adopt'. However, the noise influence on these phonemes and syllables in the voice signals is limited. By detecting the characteristics of these phonemes and syllables, directly using spoken audio can still accurately locate the related video segments. 3) Considering the actual case aiming to obtain the query results quickly after the voice input, grounding the video directly through spoken audio is intuitively more efficient than transcribing it to text through ASR.

To alleviate the abovementioned problems, we propose a video-guided curriculum learning (VGCL) strategy applied during the audio pretraining process to focus on the new task. Our method is similar to the infants' learning process when trying to understand the relationship between external visual events and spoken language. First we utilize contrastive prediction learning (CPC) \cite{DBLP:journals/corr/abs-1807-03748} to pretrain audio, which can capture long-term relations while maintaining local features from raw speech audio by employing an autoregressive model and contrastive estimation. The input training audio is mixed with random environment noise. Thus some phonemes may fail to predict due to interference. However, since the semantic information described by the video's ground truth segment is consistent with the information represented by the speech, introducing the video content affords the CPC model more external knowledge to rectify the discriminative noisy phonemes of the speech signals. During the initial training, the visual input is ground truth video segment. Then it gradually shifts to the entire video clip during the training process. With a holistic understanding of the video content first, the model can learn the corresponding relationship between actions and phonemes step by step. Extensive experiments on the ActivityNet Speech dataset show that our pretrained audio encoder can greatly advance the performance, which demonstrates the effectiveness of our proposed video-guided curriculum learning scheme.

    
    

\section{Related Work}
\subsection{Spoken-Visual Understanding}
Harwath et al.\cite{DBLP:journals/ijcv/HarwathRSCTG20} first investigate whether unsegmented raw speech audio can highlight the relevant regions in the images. This work demonstrates that this problem is quite challenging since the portions of the speech signals refer to shorter objects involving many categories. Nortje et al.\cite{DBLP:journals/corr/abs-2012-05680} investigate how to study a shared embedding space of spoken words and images with a multi-modal few-shot learning. To further explore the relationships between images and speech, Hsu et al. \cite{hsu2020text} collect a new spoken audio captions dataset from MSCOCO and introduce a model that directly generates fluent spoken audio captions of given images without any auxiliary text supervision. These works focus on the visual grounding of spoken language with static images. However, for humans to understand surrounding world, visual perception is inherently extended in time, as many spoken languages are related to the events or actions evolving over time rather than still images \cite{DBLP:journals/corr/abs-2104-13225}. Thus, recently some works have tackled this problem by focusing on the constrained domain of cooking videos \cite{DBLP:conf/cvpr/MonfortJLHFGO21, DBLP:conf/cvpr/BoggustAJHTFGZ019, DBLP:journals/corr/abs-2006-09199}. For example, \cite{DBLP:conf/cvpr/BoggustAJHTFGZ019} explore the correlations among visual images and the corresponding spoken descriptions in cooking videos with unsupervised semantic learning. Rouditchenko et al. \cite{DBLP:journals/corr/abs-2006-09199} investigate the relationship between spoken words and visual entities in videos with a contrastive loss. However, their model is trained only on instructional videos, limiting in the types of visual scenes and spoken languages. Onsescu et al. \cite{oncescu2021queryd} introduce a new speech-video retrieval dataset.
In this work, we employ the ActivityNet Speech dataset collected from ActivityNet, which is a open domain video collection and contains over 20k untrimmed videos. 

\subsection{Spoken Question Answering}
Lee\cite{lee2018odsqa} first propose a open-domain spoken question answering (SQA) dataset. Li\cite{li2018spoken} also introduce a spoken dataset for Stanford Question Answering Dataset (SQuAD). Many methods have been proposed to solve SQA problems\cite{chen2021self, you2021self, you2020contextualized, you2020towards}. \cite{you2021self} propose a temporal-alignment attention mechanism to learn cross-modality alignment between speech and text embedding spaces. \cite{you2021mrd} propose a knowledge distillation which can utilize the teacher trained on manual transcriptions to guide the training of the student on ASR transcriptions.

\subsection{Video Grounding}
Text-to-video temporal grounding is first studied in \cite{DBLP:conf/iccv/HendricksWSSDR17, DBLP:conf/iccv/GaoSYN17}, aiming to localize the time interval in a video semantically relevant to the text description. Various models have been proposed to solve this problem, which can be divided into proposal-based and proposal-free methods. Proposal-based methods first extract candidate proposals with sliding windows and then rank the candidates with text query \cite{DBLP:conf/aaai/ChenJ19a, DBLP:conf/cvpr/ZhangDWWD19, DBLP:conf/sigir/ZhangLZX19}. Zhang et al. \cite{DBLP:conf/cvpr/ZhangDWWD19} construct a graph to explicitly model moment-wise temporal relations. \cite{DBLP:conf/sigir/ZhangLZX19} learn fine-grained representation learning by devising a syntactic GCN to leverage the syntactic structure of queries and apply multi-head attention to capture long range semantic dependencies from video context. \cite{ma2021hierarchical} use convolutional neural network to generate coarse video candidate moments and then adjust the boundary. \cite{hu2021coarse} design a hierarchical semantic tree to explore the semantic relationship between different visual moments. \cite{DBLP:journals/corr/abs-2008-02448} propose a fine-grained iterative attention module to incorporate information from text and video mutually. Although these methods have achieved promising results, they are suffering from time-consuming and computation-extensive problems. 

Recently, thanks to the development of machine reading comprehension in NLP, proposal-free models have been proposed to alleviate the above mentioned problems. For example, \cite{DBLP:conf/aaai/Chen0CJL19} use a cross-gated attended recurrent module to exploit the interactions between text query and video, and then use a segment localizer to directly predict start and end moment. \cite{DBLP:conf/naacl/GhoshAPH19} apply an extractive approach which can leverage cross-modal interactions between text and video to predict the start and end frames. \cite{zhang2020span} propose a simple but effective query-guided highlighting (QGH) strategy which can guide the model to search for matching video span within a highlighted region.

\subsection{Curriculum Learning}
Inspired by human learning process, Bengio et al. \cite{DBLP:conf/icml/BengioLCW09} propose curriculum leaning that gradually increased the training data complexity. Many studies follow the original formulation that applies curriculum strategy at the data level \cite{DBLP:journals/csl/ShiLJ15, DBLP:conf/cvpr/PentinaSL15}. There are also some methods that apply curriculum strategy at the model level, which gradually increase the model's capacity during the training process \cite{DBLP:conf/iclr/KarrasALL18, DBLP:conf/nips/SinhaGL20}. Other works explore the curriculum learning at the task level. \cite{DBLP:conf/iccvw/SarafianosGNK17} first group the tasks into strongly and weakly-correlated ones and then transfer the knowledge from the strongly-correlated tasks to the weakly-correlated tasks. 
Jiang et al. \cite{NIPS2014_c60d060b} first propose a self-paced learning with diversity methodology, which formalizes the preference for both easy and diverse samples into a general regularizer. \cite{DBLP:conf/ijcai/LiangJMH16} combine the curriculum learning and self-paced learning together for webly labeled video data learning with noisy labels. \cite{DBLP:conf/naacl/PlataniosSNPM19} propose a continuous curriculum learning in neural machine translation. In this paper, we propose a curriculum learning strategy similar to the data level methods, which gradually shift the given video content from ground truth part to the entire part during pretraining audio encoder utilizing contrastive learning. 

\section{ActivityNet Speech Dataset}
To investigate whether machine can associate the unsegmented raw spoken language with related video contents together just like human being, we first construct a large-scale spoken video grounding dataset based on ActivityNet Caption \cite{DBLP:conf/iccv/KrishnaHRFN17} and name it as ActivityNet Speech dataset. The new dataset contains over 70k speech annotations and 20k videos. The videos are open-domain and contain many kinds of activities. Each video is annotated with 3.65 natural language descriptions on average and marked with aligned temporal boundary timestamps and the average duration of videos is 117.74 seconds. To obtain the speech annotations, we employ 58 speakers to read the original text descriptions, which contain 28 male speakers and 30 female speakers. We try to keep the balance between different genders in order to avoid this becoming a confounding factor interfering with the model reasoning process. To guarantee the recording quality, we ask all the speakers to read smoothly without a stammer.
The average of each speech recording is 6.22 seconds and about 124.3 hours in total. We split the dataset into 37417, 17505 and 17031 clip-audio pairs for training, validation, and testing, respectively.

\begin{table}[]
\small

\caption{The statistics of the Activity Speech dataset}
   \begin{center}

   \begin{tabular}{c|ccc}
   \hline
   
   \textbf{Split}  & \textbf{train}   & \textbf{val} & \textbf{test} \\ \hline \hline

Number of items  & 37421 & 17505 & 17031 \\
Avg. length of audio & 6.36s & 6.38s & 5.74s \\

\hline
   \end{tabular}
   \end{center}

\label{Statistic}
\vspace{-2.0em}
\end{table}         

\section{Method} \label{main model}
In this section, we first introduce the novel spoken video grounding task; then we give a description of the overall architecture of our proposed network; finally we detail the pre-training process of the audio CPC with video-guided curriculum learning (VGCL) strategy.

\subsection{Spoken Language Video Grounding Task}
Given a spoken audio $A$ and its corresponding video clip $V$, the goal of the spoken language video grounding task is to localize the start and end video frames that semantically meet the description of the audio. In order to simulate the real-world scenarios, we randomly select an environmental sound \cite{10.1145/2733373.2806390} that we consider as noise and merge it with the clean speech at a random ratio. Compared with traditional text-to-video grounding, the new task is far more challenging due to the unsegmented and unaligned noisy spoken audio. To bridge this gap, we propose a novel curriculum learning strategy that gradually understands the noisy spoken language with corresponding video content. Essentially, our method can imitate the process of babies trying to understand the relationship between spoken language and external visual events.

\subsection{Overall Architecture}
\begin{figure}[tb] 
 \center{\includegraphics[width=8cm]  {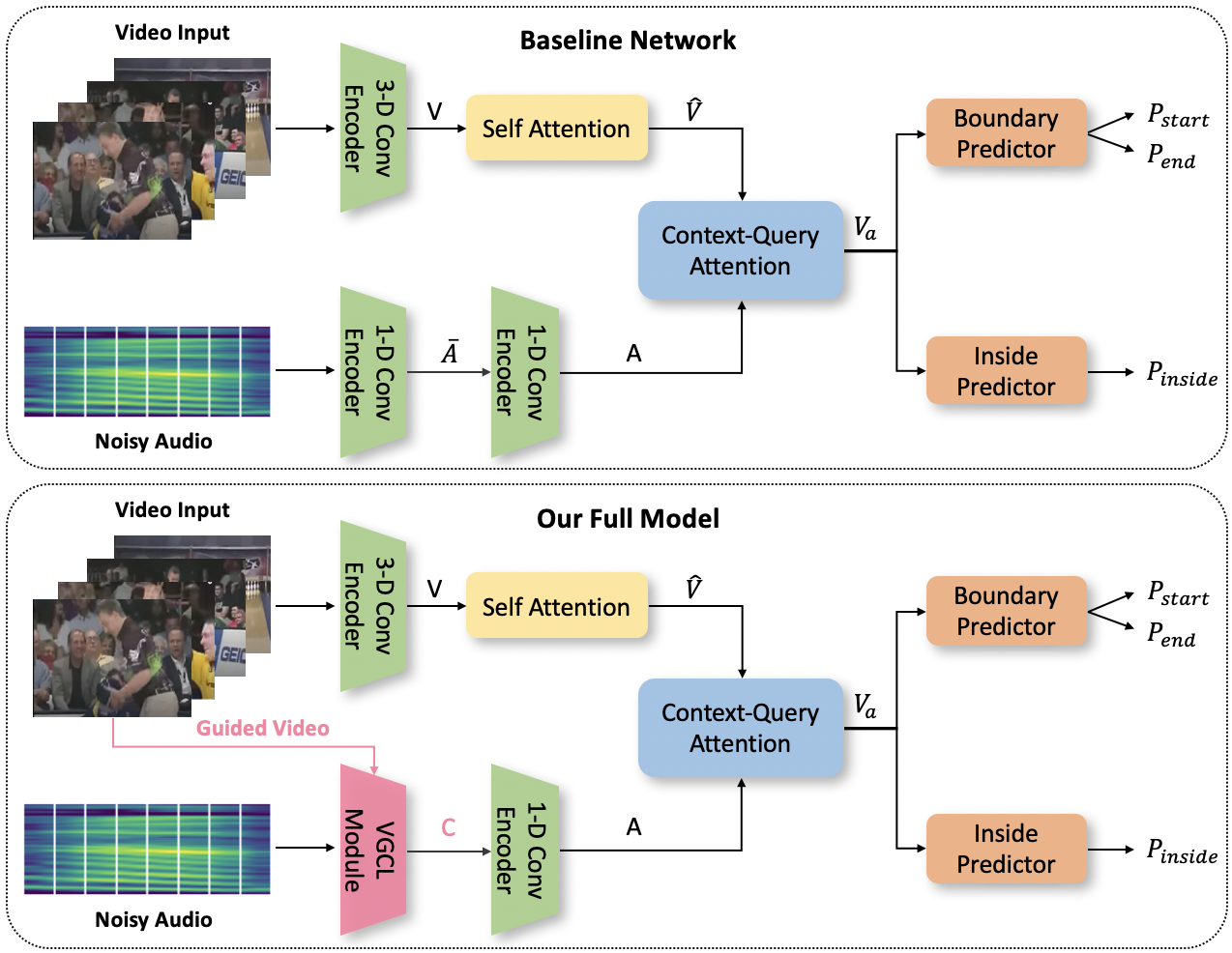}} 
 \caption{\label{fig2} The overview of our proposed network. In our full model, we use the pretrained video-guided curriculum learning module to replace the first audio 1D convolution encoder.}
 \vspace{-0.7cm}
\end{figure}

As depicted in Fig \ref{fig2}, our baseline model can be divided into four modules. Specifically, after encoding the video and spoken language into the same latent space, we can obtain the visual features as $V = \{v_{i}\}_{i=1}^{N_{v}}$, and the audio features as $A = \{a_{i}\}_{i=1}^{N_{a}}$, where $N_{v}$ and $N_{a}$ are the number of the extracted visual and audio frames, respectively. 
Then, we apply a self-attention layer to obtain a holistic understanding of the video content and utilize a context-query attention module to integrate the visual and audio features. Finally, we employ a prediction layer based on recurrent network to predict the temporal grounding result. Our main contribution is the audio pre-training process based on the video-guided curriculum learning. After obtaining the audio pre-trained model, we exploit it to replace the audio encoder in feature extraction process. More details about the curriculum learning strategy are presented in Section \ref{VGCL}. 

\subsection{Encoders} \label{encoder}
\textbf{Audio encoder} We represent the noisy audio input as a log Mel filterbank spectrogram, with a 16 kHz sampling rate, a 25 ms Hamming window, a 10 ms window stride, and 128 Mel filter bands. We apply two 1-D convolution blocks and a single layer unidirectional GRU to encode the spectrogram. Then we use two trainable CNN blocks with residual layers \cite{DBLP:journals/ijcv/HarwathRSCTG20} to process the audio features into $A = \{a_{i}\}_{i=1}^{N_{a}} \in R^{N_{a} \times d}$, where d denotes the feature dimension. 

\textbf{Video encoder} For the untrimmed visual features extracted by pretrained C3D model\cite{tran2015learning}, we use a 1D convolutional layer and one layer bi-directional GRU layer as our visual encoder, which can capture a long range information. We encode the visual features into the same dimension latent space with audio features as $V = \{v_{i}\}_{i=1}^{N_{v}} \in R^{N_{v} \times d}$. 

\subsection{Visual-Audio Attention and Predict Layer}
Following the previous machine reading comprehension works, we formulate the interaction between video context and audio query from the perspective of Context-Query Attention (CQA) module, which have been successfully used in many text-to-video grounding methods \cite{zhang2020span, DBLP:conf/cvpr/ZhaoZZL21}. We first input the visual features into a self-attention layer to get a holistic understanding of the entire video content: $\hat{V} = SelfAtt(V)$. Then we use the CQA to calculate the cross-modal similarity score $S \in R^{N_{v}\times N_{a}}$ between the visual and audio features. The video-to-audio attention $\beta_{1} \in R^{N_{v} \times d}$ and audio-to-video attention $\beta_{2} \in R^{N_{v} \times d}$ can be obtained by:
\begin{equation}
    \beta_{1} = S_{r} \cdot {A}, \quad \beta_{2} = S_{r} \cdot  S_{c}^{T} \cdot \hat{V},
\end{equation}
where $S_{r}$ and $S_{c}$ are the normalized result of similarity score $S$ by SoftMax along row and column axis, respectively. Finally these two attention matrices are fused to integrate the visual-audio attention vectors:
\begin{equation}
    V_{a} = FFN([\hat{V}; \beta_{1};\hat{V}\odot \beta_{1};\hat{V}\odot \beta_{2}] \in R^{N_{v}\times d},
\end{equation}
where FFN is the feed-forward layer and $\odot$ is the element-wise multiplication.

We use two unidirectional GRU to predict start and end boundaries $F_{s}$ and $ F_{e}$. Inspired by \cite{zhang2020span}, the end boundary probability is calculated based on the start boundary probability, since an event's start and end boundaries are always related:
\begin{equation}
    F_{s} = \overrightarrow{GRU_{s}}(V_{a}), \quad
    F_{e} = \overrightarrow{GRU_{e}}(F_{s}),
\end{equation}

Instead of solely relying on sophisticated start and end boundaries prediction, we introduce an additional unidirectional GRU to predict inside frame probability $F_{i}$, helping the model distinguish whether a frame belongs to an event. Then we apply three feed-forward layers to predict the boundary and the inside frame probability scores:
\begin{gather}
    P_{s} = FFN([F_{s};V_{a}]),
    P_{e} = FFN([F_{e};V_{a}]),\\
    P_{i} = FFN([F_{i};V_{a}]),
\end{gather}
where $P_{s}$ and $P_{e}$ denote scores of the start and end boundaries at each position, $P_{i}$ denotes the scores of the inside frames whether relate to the spoken language description. 

\begin{figure*}[htb] 
 \center{\includegraphics[width=18cm]  {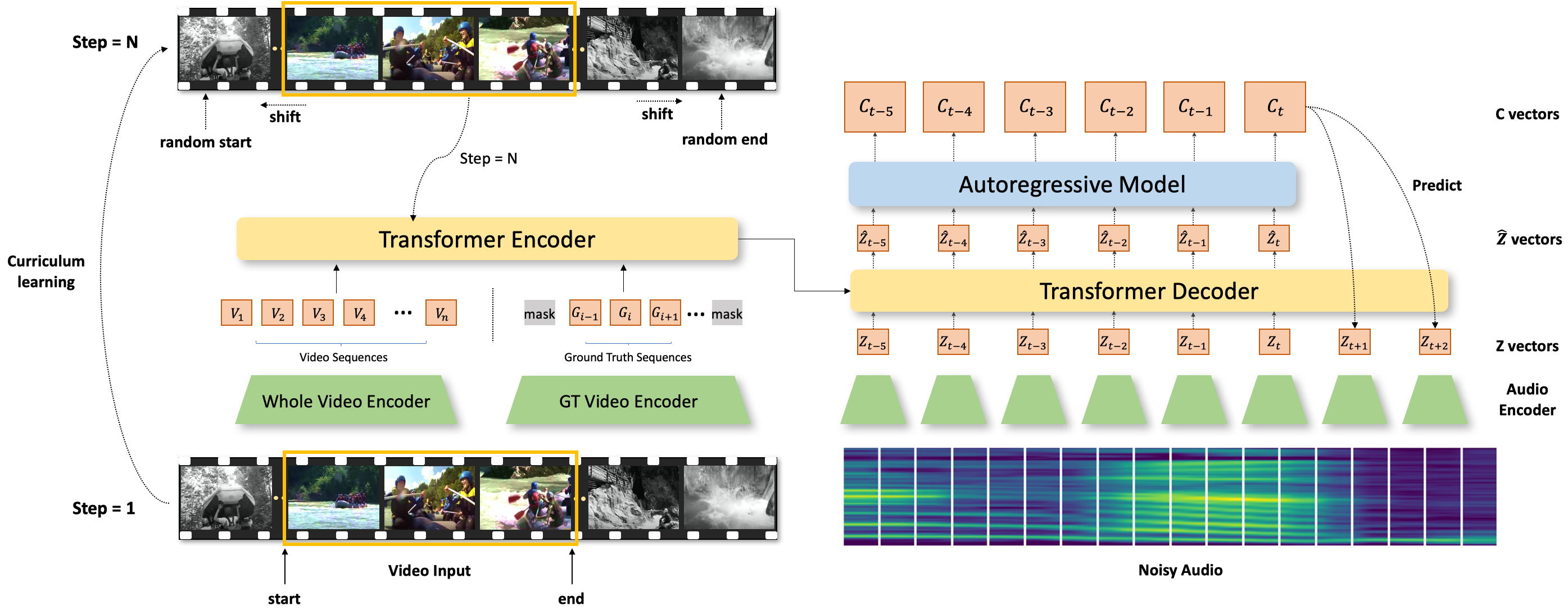}} 
 \caption{\label{fig3} The proposed video-guided curriculum learning module. (1) We extract the ground truth video features and mask the remaining parts during the initial training. Then we concatenate it with the entire video to obtain a comprehensive understanding. We process the video input with a transformer encoder layer. (2) We set the output of the visual features as the key and value matrices and set the audio features as the query. Then we employ a cross attention mechanism to extract the visual information that is related to audio. (3) The input video is gradually shifted from the ground truth part to the entire video content during the training process. Finally, the model can learn how to extract semantically related information from the entire video clip, distinguishing the ambiguous speech phonemes. } 
\end{figure*}

\section{Denoise Spoken Audio with Video-Guided Curriculum Learning } \label{VGCL}
In this section, we will illustrate our proposed VGCL for the audio pretrain process. First, we introduce how to pretrain the audio spectrogram encoder by contrastive predictive coding (CPC) \cite{DBLP:journals/corr/abs-1807-03748}. Then we introduce the purpose of using video information to assist audio learning, and how to gradually shift the given video content from ground truth part to the entire video part. Finally we detail three different pacing functions for curriculum learning schedule.

\subsection{Audio pretrain with CPC} \label{CPC}
CPC has been widely used in audio pre-training, which can extract high-level phonetic representations and acoustic units from raw observations by predicting the future samples with powerful autoregressive models. However, most of the previous works predict the future audio waveform features, which is not suitable for our task. Thus in this paper, we propose to directly train CPC on audio log-Mel spectrogram $\tilde{A}$. The spectrogram is first processed by a stack of 2 convolutional layers as encoder $g_{enc}$, each downsamples the input by a factor of 2. Thus we can get a sequence of latent variables $z_{t} = g_{enc}(\tilde{A}_{t})$. Then we use a single layer undirectional GRU to summarize the information of all $z_{\leq t}$ and obtain a context representation as $c_{t} = GRU(z_{\leq t})$. 

Given a prediction of $K$ steps, a set $Z$ of $N-1$ random negative samples and one positive sample $z_{t+k}$, we use $c_{t}$ to predict the k-th future step $z_{t+k}$, and the Info Loss can be optimized as:
\begin{equation}
    L_{K} = -\frac{1}{K}\sum_{k=1}^{K}log[\frac{exp(z_{t+k}^{T}W_{k}c_{t})}{\sum_{z_{j} \in Z}exp(z_{j}^{T}W_{k}c_{t})}],
\end{equation}
where $W_{k}$ is the linear projection matrix for different step k. After pretraining, we can use the encoder $g_{enc}$ and GRU layer to replace the according audio encoder layer in Section \ref{encoder}. 

\subsection{Video-Guided Curriculum Learning}

\textbf{Motivation:} We all have such experiences, when the speech is disturbed by external noise, it is easy for us to misunderstand some keywords, i.e. "sing" or "seen" as shown in Fig \ref{fig4}. Such mistakes will lead us to completely misunderstand the original sentences. However, if we see the corresponding events while hearing the noise speech, the visual information will help us to rectify these errors. Besides, when babies first learn to understand the relationship between noisy speech and visual actions in the external world, their parents will point out the related visual events while repeating the descriptive speech. The introduction of visual information can help babies learn how to distinguish between noise and real useful voice information progressively.

Inspired by this, in this section, we propose a video-guided curriculum learning strategy, which can make use of visual perception to help audio pretrain with CPC and effectively suppress the environmental noise. The overall architecture of VGCL is shown in Figure \ref{fig3}. Specifically, given a piece of audio, we first use self-attention to have a comprehensive understanding of its corresponding video content as $\tilde{V}$, then we use mask mechanism to keep the critical part of the video content. Concretely, the left and right mask boundary $M_{l}$ and $M_{r}$ can be calculated as follow:
\begin{equation}
    M_{l} = \tau_{s}-\frac{t}{\kappa} \times \gamma(\tau_{s} - 0), \quad M_{r} = \tau_{e} + \frac{t}{\kappa} \times \gamma(L - \tau_{e})
\end{equation}
where $L$ is the video length, $\tau_{s}$ and $\tau_{e}$ are the ground truth video start and end points, $\gamma \sim U(0, 1)$, $\kappa$ is the number of the curriculum learning stages, $t \in (0, \kappa)$. We use the mask matrix to obtain the masked video content as $\tilde{V}_{mask}$.

In the initial training stage, the visual input is the ground truth video part. Then we continue training the model with t increasing sequentially. For simplicity, we increase t from 0 to $\kappa$ step by step, which means the input video part is gradually shifting from ground truth to the entire fragment. 
By understanding the information in this vital video content, we can learn the connections between phonemes and corresponding actions in the video, retaining key features and suppressing environmental noise during CPC autoregressive encoding. Meanwhile, we also find in the experiments that the effectiveness of solely adding ground truth video part is not satisfying due to the lack of video context information. Therefore, we concatenate the entire video clip with the masked video content in the time dimension: $\tilde{V}_{full} = [\tilde{V}; \tilde{V}_{mask}] $. 

Then we project audio into the query features $Q_{a}$, and project the concatenated visual features into key and value features $K_{v}$ and $V_{v}$. Finally we apply cross-modal attention to exploit related information from visual features:
\begin{equation}
    \hat{Z} = LayerNorm(\tilde{A} + \delta (Softmax(Q_{a} K_{v}^{T})V_{v})),
\end{equation}
where $\delta$ denotes the Relu function. Then we use $\hat{z}$ to calculate $\hat{c}_{t} = GRU(\hat{z}_{\leq t})$, but still use $\hat{c}_{t}$ to predict $z_{t+k}$. The Info loss can be optimized as:
\begin{equation}
    \hat{L}_{K} = -\frac{1}{K}\sum_{k=1}^{K}log[\frac{exp(z_{t+k}^{T}\hat{W}_{k}\hat{c}_{t})}{\sum_{z_{j} \in Z}exp(z_{j}^{T}\hat{W}_{k}\hat{c}_{t})}],
\end{equation}
where $\hat{W}_{k}$ is the linear projection matrix for different step k. After the pre-training process, we can obtain a mature audio encoder, which can extract vital visual information from the video clip and use it to rectify ambiguous phonemes from noisy speech input. We use the VGCL module to replace the first audio 1D convolution encoder as described in Fig \ref{fig2}. The experiments demonstrate that with the addition of VGCL module, our full model can effectively suppress the noise and improve the localization performance compared with baseline network. 

\subsection{Pacing Functions} \label{pacing functions}
Pacing function is widely used in curriculum learning strategies, which can control the training steps for different training stages. Concretely, we divide our training process into $\kappa$ stages, and define three different pacing functions to control the training steps as $t \in (0, \kappa)$ increases: linear, exponential, and logarithmic \cite{DBLP:conf/ijcai/LiuRTZQZL20}. Linear function refers to exploiting the same training steps for each stage. Exponential function means that for early t, the training steps are small and become more prominent as the input video boundaries get blurred. Logarithmic function is entirely different from the exponential function as the training steps will gradually decrease as the t increases. We compare and analyze these different pacing functions with experiments to demonstrate the impact of different curriculum learning strategies on audio pretraining.

\subsection{Training and Inference}
Based on the abovementioned audio pretrain process with video-guided curriculum learning and main model structure, we can get the start and end boundary probability scores: $P_{s}$ and $P_{e}$, where the probability distributions can be calculated as $\hat{P_{m}} = SoftMax(P_{m})$,  $m \in \{s, e\}$; and one inside frame probability scores: $P_{i}$, where the distributions can be calculated as $\hat{P_{i}} = Sigmoid(P_{i})$. Then we use these probability distributions to define two loss functions, boundary loss and inside loss:

\textbf{Boundary loss:} We utilize cross-entropy loss function to optimize the start and end boundary predictions, the loss item can be defined as:
    \begin{equation}
        L_{bound} = \frac{1}{2} [CE(\hat{P_{s}}, \tau_{s}) + CE(\hat{P_{e}}, \tau_{e})]
    \end{equation}
    
\textbf{Inside loss:} We also introduce the inside frame loss to let the model distinguish whether a frame belongs to answer or not, which can further optimize the boundary prediction. Specifically, the inside loss item can be defined as:
    \begin{equation}
        L_{in} = -\sum_{t=1}^{N_{v}}[\hat{P_{i}^{t}}\log y_{t} + (1-\hat{P_{i}^{t}})\log(1-y_{t})]
    \end{equation}
    where $y_{t}$ is the ground truth label for each frame, if the t-th frame belongs to the answer fragment, then the value of $y_{t}$ is 1, otherwise is 0.

The overall loss function of the main model can be summarized as $L_{total} = L_{bound} + L_{in}$.

At the inference time, the start and end boundary predictions can be calculated by maximize the probability scores $P_{s}$ and $P_{e}$.

\section{Experiments}


\subsection{Data Processing}
For visual features, we follow the previous works \cite{zhang2020span, DBLP:conf/cvpr/MunCH20} and use the features extracted by a publicly available pre-trained C3D model \cite{tran2015learning}. For each speech audio, we first randomly select a piece of noise from ESC-50 \cite{10.1145/2733373.2806390}, and then add the noise to a random part of the original audio with a proportion $\alpha \in (0.5, 0.7)$. ESC-50 dataset is a labeled collection of 2000 environmental audio and consists of 5-second-long recordings organized into 50 semantical classes, which can be divided into five main classes: animals, natural soundscapes/water sounds, human non-speech sounds, interior/domestic sounds and exterior/urban noises.

We downsample the noisy audio recordings to 16KHz and extract 128-dim mel-spectrogram features. For the convenience of model training, we uniformly sample segments from each mel-spectrogram feature with a fixed length $N_{a} = 1024$. 
We use the Deepspeech \cite{ppspeech2021} to translate the noise speech audio into text and obtain the word embeddings with the 300d Glove model \cite{DBLP:conf/emnlp/PenningtonSM14}. 

\subsection{Model Configurations and Evaluation Metrics}
We set the frame number of video $N_{v}$ as 64, and the frame length of audio $N_{a}$ as 1024 for ActivityNet Speech dataset. The hidden size of all layers for both audio and visual features is set to 1024. We split the features into 8 chunks to improve the model stability, and average the results of all chunks at last. The prediction step K for audio pretraining process is set to 3, and the default audio length is set to 256. For curriculum learning, the default pacing function is linear function, and $\kappa = 10$. During training, we apply Adam optimizer to optimize the model with a warmup strategy, the initial learning rate is set to 0.001. Following the previous works \cite{DBLP:conf/iccv/GaoSYN17, DBLP:conf/cvpr/MunCH20}, we apply "R@n, IoU = m" and "mIoU" as our evaluation metrics.


\begin{table}[t]
\small
\caption{Comparisons with baselines on ActivityNet Speech dataset,  n = 1 and $m \in \{0.3, 0.5, 0.7\}$}
   \begin{center}

   \setlength{\tabcolsep}{1mm}
   \begin{tabular}{c|l|ccc|c}
   \hline
   
   \textbf{Query} &\textbf{Models}  & \textbf{IoU=0.3}   & \textbf{IoU=0.5} & \textbf{IoU=0.7}  & \textbf{mIoU} \\ \hline \hline

Audio&Base        & 44.93 & 28.59 & 15.40 & 31.89 \\
Audio&Base+VQCPC  & 44.29 & 26.18 & 14.90 & 31.94 \\
Audio&VSLBase \cite{zhang2020span}    & 46.12 & 29.10 & 15.89 & 33.20 \\
Audio&Base+CPC    & 47.42 & 28.78 & 16.12 & 33.83 \\
Audio&VSLNet \cite{zhang2020span}    & 46.75 & 29.08 & 16.24 & 34.01 \\
Audio&Full model  & \textbf{49.80} & \textbf{30.05} & \textbf{16.63} & \textbf{35.36} \\
\hline
\hline
Text&Base+text(ASR) & 48.49 & 29.90 & 16.33 & 34.58 \\
Text& -Glove        & 50.46 & 27.32 & 15.83 & 33.82 \\
Text&Base+text(Ori) & 60.57 & 42.96 & 25.68 & 43.34 \\
\hline
   \end{tabular}
   \end{center}

\label{main result}
\vspace{-2.0em}
\end{table}         

\subsection{Compared with Baseline Model}
Since the spoken video grounding is a new task, none of the previous works focus on handling it. Therefore, we set the model as depicted in Section \ref{main model} as our baseline model, which directly uses raw speech to localize the video fragments without any audio pretrain process, also we implement VSLNet \cite{zhang2020span} with audio by ourselves. Besides, we compare our proposed method with another audio pretrain model, VQCPC \cite{DBLP:journals/corr/abs-2106-10132}, which combine VQ-VAE \cite{DBLP:conf/nips/OordVK17} and CPC together to extract speech representations. We also use the DeepSpeech to translate the noise audio into text to test the performance of the baseline model with these ASR transcripts. 

The main results are shown in Table \ref{main result}. Specifically, our proposed method can outperform the baseline model with a large margin, especially achieving score 1.23$\%$ absolute improvements in $IoU = 0.7$. The results of row 2 and row 4 indicate that the vanilla CPC is more suitable for this task. Further, as shown in row 4 and row 6, with the introduction of the video content, the performance is improved with a notable margin in both IoU and mIoU scores, which demonstrates that the video information can facilitate the audio denoising. The result in row 7 and row 8 indicate that although the text transcripts recognized by ASR have lots of errors, the introduction of Glove word embedding model can still guarantee video grounding performance. However, with the equipment of the proposed curriculum learning strategy, our VGCL can surpass the text-based grounding model in both accuracy and efficiency.

\subsection{Ablation Studies}

\textbf{Model components} To investigate the effect of individual components in our proposed VGCL module, we conduct a series of ablation studies, the results are shown in Table \ref{Ablation 1}. To demonstrate the effectiveness of the curriculum learning, we remove it and simply use the entire video content to do cross-attention with audio features, denotes as "w/o CL". The scores drop a lot, which shows that without the highlight of ground truth visual features, the model can barely extract useful information from complex video contents. To further verify the effectiveness of the introduction of the video, we remove the entire video content, leaving only the masked video part, which denotes "w/o entire video". Also we remove the video self-attention layer, which denotes as "w/o self attention". The results show that both the entire video content and the self-attention layer can help extract context visual information, which is vital for audio pretraining. The default audio length when training CPC is set to 256, here we test the performance when the audio length is set to 128. The score decreases 0.3 in $IoU = 0.7$, indicating that the suitable length will benefit the audio pretraining process, short audio segments are difficult to extract relative information from the input video. 
\begin{table}[t]
\small

\caption{Ablation study on the effect of different modules or settings in our video-guided curriculum learning (VGCL).}
   \begin{center}

   \begin{tabular}{l|ccc|c}
   \hline
   
   \textbf{Models}  & \textbf{IoU=0.3}   & \textbf{IoU=0.5} & \textbf{IoU=0.7}  & \textbf{mIoU} \\ \hline \hline

full model  & \textbf{49.80} & 30.05 & \textbf{16.63} & \textbf{35.36} \\
w/o CL             & 46.83 & 29.19 & 16.08 & 33.04 \\
w/o entire video    & 48.79 & 29.92 & 16.28 & 33.94 \\
w/o self attention & 49.00 & 30.09 & 16.42 & 34.65 \\
w/o random         & 48.12 & 30.12 & 16.59 & 34.20 \\
audio length=128   & 48.97 & \textbf{30.33} & 16.29 & 34.44 \\
\hline
   \end{tabular}
   \end{center}

\label{Ablation 1}
\end{table}         

\begin{table}[t]
\small

\caption{Ablation study on the effect of different pacing functions and different curriculum learning steps $\kappa$. }
   \begin{center}

   \setlength{\tabcolsep}{1mm}
   \begin{tabular}{l|ccc|c}
   \hline
   
   \textbf{Pacing Functions}  & \textbf{IoU=0.3}   & \textbf{IoU=0.5} & \textbf{IoU=0.7}  & \textbf{mIoU} \\ \hline \hline

Linear ($\kappa = 10$) & \textbf{49.80} & {30.05} & \textbf{16.63} & \textbf{35.36}\\
Linear ($\kappa = 5$)   & 48.82 & 30.03 & 16.25 & 34.44 \\
Exponential ($\kappa = 10$)     & 49.46 & \textbf{30.34} & 16.57 & 34.96 \\
Exponential ($\kappa = 5$)     & 48.34 & 29.59 & 16.39 & 34.44 \\
Logarithmic ($\kappa = 10$)    & 48.89 & 29.98 & 16.40 & 34.53 \\
Logarithmic ($\kappa = 5$)    & 46.61 & 29.06 & 16.21 & 33.30 \\

\hline
   \end{tabular}
   \end{center}
\label{Ablation 2}
\end{table}         

\begin{figure}[htb] 
 \center{\includegraphics[width=8cm]  {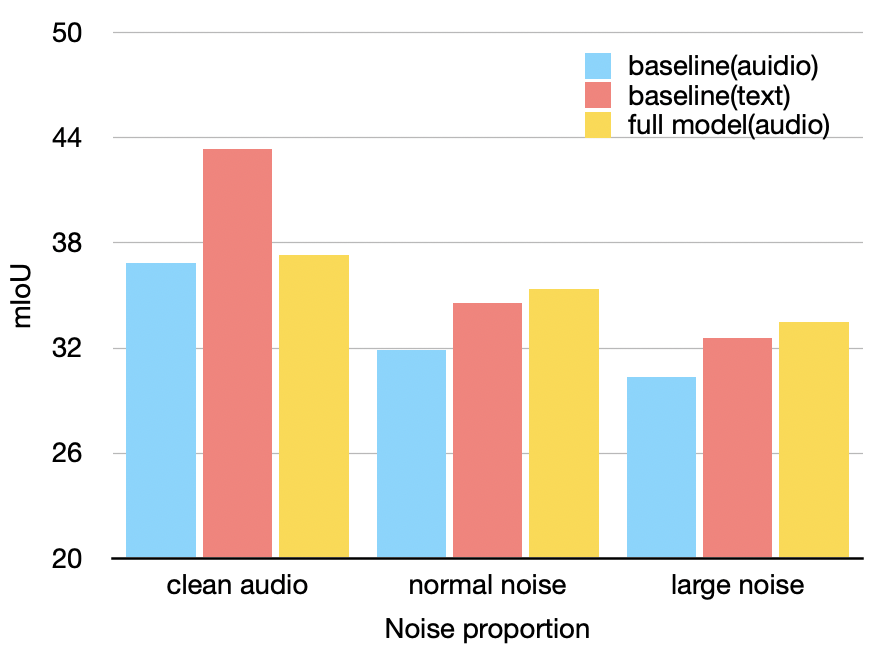}} 
 \caption{\label{noise} Analysis of the impact of noise proportion $\beta$. For clean, normal, large noise, $\alpha \in $  0, (0.5, 0.7), (0.7, 0.9), respectively. For normal and large noise situation, the text is the results translated by DeepSpeech.} 
\vspace{-0.5em}
\end{figure}

\begin{figure}[htb] 
 \center{\includegraphics[width=8cm]  {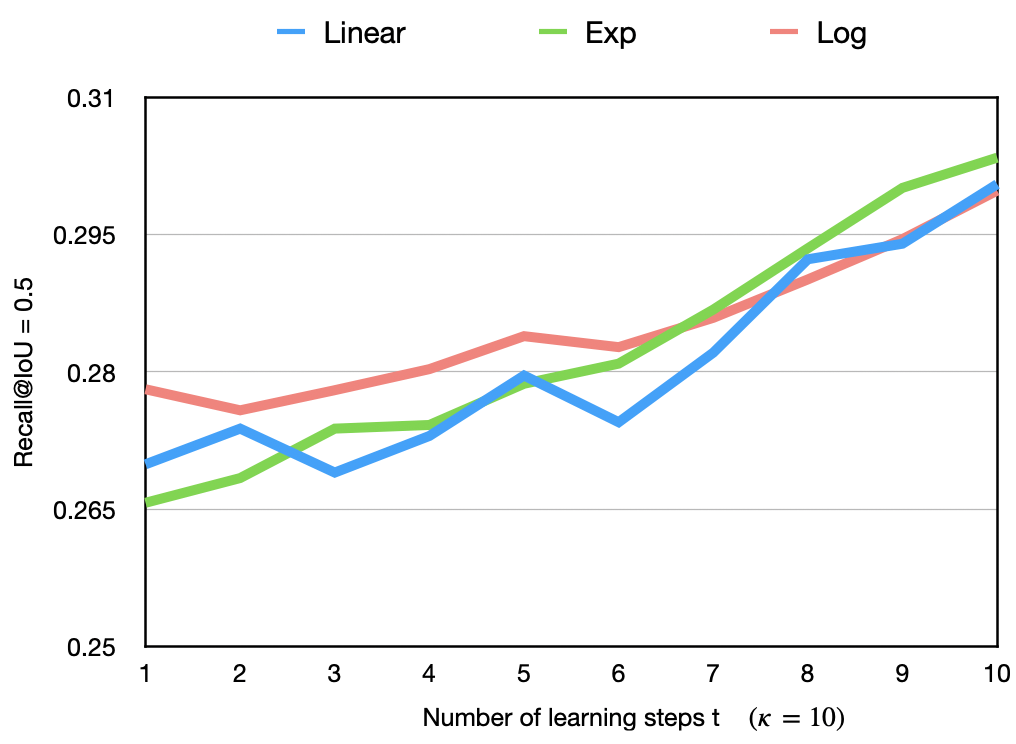}} 
 \caption{\label{curriculum} Analysis of different strategies for curriculum learning.} 
 \vspace{-0.5em}
\end{figure}

\begin{figure*}[htb] 
 \center{\includegraphics[width=16cm]  {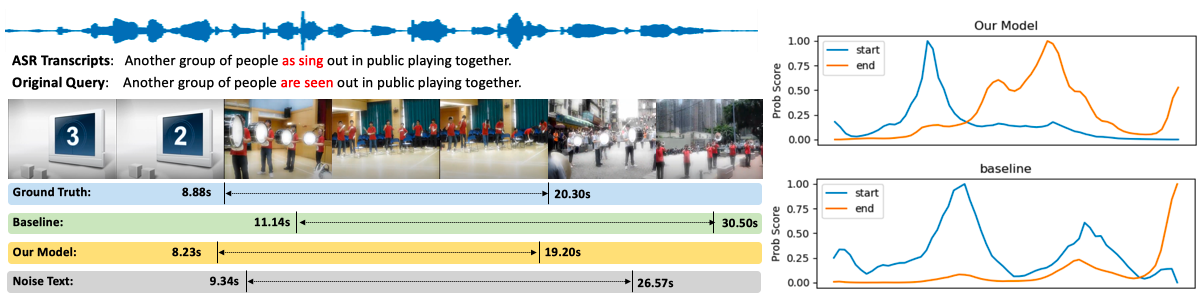}} 
 \caption{\label{fig4} Success case of our model on ActivityNet Speech dataset.} 
\end{figure*}

\begin{figure*}[htb] 
 \center{\includegraphics[width=16cm]  {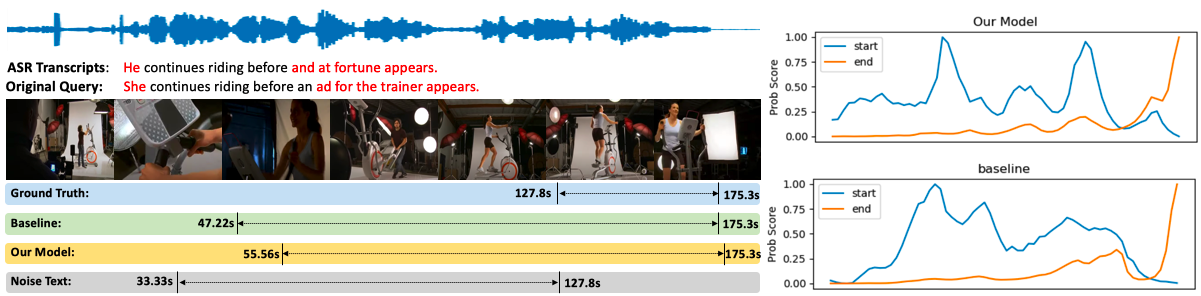}} 
 \caption{\label{fig5} Failure case of our model on ActivityNet Speech dataset.} 
\end{figure*}

\textbf{Analysis on Pacing Functions}
We conduct a series of experiments to compare the models pretrained with different pacing functions and different curriculum learning steps. From the Table \ref{Ablation 2} we can see that the performances of the models trained with linear function and exponential function are equivalent, and slightly outperforms the model trained with logarithmic function. As we described in Section \ref{pacing functions}, logarithmic function focuses more on the easier samples while exponential function focuses more on harder samples. 
In addition, in order to explore the impact of different training steps in the pre-training stage, we test the effect of the pretrained audio encoder when $t \in (1, \kappa)$ with different pacing functions, as shown in Fig \ref{curriculum}. The results show that logarithmic function performs better when t is small, and exponential function achieves the best performance when the pre-training is finished. These also demonstrate that more training steps on harder curriculum stages is beneficial to our task.

\textbf{Analysis on Different Proportion of Noise}
To further verify the effectiveness of our model, we compare the performances of the models trained under different proportions of noise, as shown in Fig \ref{noise}. The results demonstrate that under different proportions of noise, our VGCL can facilitate the contrastive prediction process and obtain a mature audio encoder for spoken video grounding tasks. It is noticeable that the improvement after changing the model from baseline to our VGCL when the audio is clean is lower than the situation where the audio is noisy. From the figure we can also find that when the audio is clean, the text-based method performs better. However, when the proportion of noise is getting larger, our model can surpass the text-based method by a certain margin. 
This indicates that our VGCL can utilize the corresponding video content to rectify the noisy phonemes and improve the grounding performance, which is superior in real application.

\subsection{Qualitative Analysis}
To further qualitatively compare our method with baseline model and text-to-video grounding model based on ASR transcripts, we analyze two examples from Activitynet Speech dataset, as shown in Fig \ref{fig4} and Fig \ref{fig5}. From the success case in Fig \ref{fig4} we can see that compared with the other two models, our VGCL can accurately locate the correct event boundaries. The ASR transcripts have a large deviation in the key verbs due to the interference of noise, leading to the wrong localized moments. This demonstrates the difficulty for the model to learn a reliable correspondence between ASR transcripts from noisy audio and video actions.

From the failure case in Fig \ref{fig5} we can observe that a woman is riding a machine in many frames, however, the key world "continue" emphasizes that the model should focus on the latter frames. Our model fails to predict the ground truth action boundaries, but still have a high score for the right start position. This example indicates that our model still lacks of the ability for understanding some key adjectives and adverbs, which needs to be strengthened in future.


\section{Conclusion}
In this paper, we introduce a new task, spoken video grounding (SVG), which aims to localize the target video segments with the corresponding speech audio. Considering this is a novel task and has no corresponding dataset, we collect a new large-scale spoken video grounding dataset based on ActivityNet Caption. To further simulate the real application, we randomly add different environment noises to these speech audio. We find the ground truth content of the video is consistent with the information described by the speech. Thus we can utilize the corresponding video information as external knowledge to rectify these discriminative noisy audio phonemes. Therefore, we bring up a video guided curriculum learning (VGCL) strategy to facilitate the audio pretrain process. Our VGCL can gradually shift the input video content from the ground truth part to the entire video clip, which makes the pretraining process from easy to hard. Finally, the model can learn how to extract critical information from the input video clips that are related to the audio description. 


\begin{acks}
This work was supported by the National Natural Science Foundation of China under Grant No.62072397, Zhejiang Natural Science Foundation under Grant LR19F020006.
\end{acks}

\bibliographystyle{ACM-Reference-Format}
\bibliography{VGCL}

\appendix

\end{document}